\documentclass[11pt,a4paper]{article}
\usepackage[hyperref]{acl2019}
\usepackage{times}
\usepackage{latexsym}

\usepackage{url}
\usepackage{soul}
\usepackage[utf8]{inputenc}
\usepackage{amsmath}
\usepackage{CJK}
\usepackage{amsfonts}
\usepackage{multirow}
\usepackage{graphicx}  
\usepackage{array}
\usepackage{enumitem}
\usepackage{algorithm}
\usepackage{algorithmic}
\aclfinalcopy 

\newcommand{\algorithmicbreak}{\textbf{break}}
\newcommand{\BREAK}{\STATE \algorithmicbreak}

\usepackage{enumitem} \setenumerate[1]{itemsep=0pt,partopsep=0pt,parsep=\parskip,topsep=5pt} \setitemize[1]{itemsep=0pt,partopsep=0pt,parsep=\parskip,topsep=5pt} \setdescription{itemsep=0pt,partopsep=0pt,parsep=\parskip,topsep=5pt}

\title{Open Vocabulary Learning for Neural Chinese Pinyin IME}

\author{
	Zhuosheng Zhang$^{1,2}$,
	Yafang Huang$^{1,2}$,
	Hai Zhao$^{1,2,}$\thanks{$\ $ Corresponding author. This paper was partially supported by National Key Research and Development Program of China (No. 2017YFB0304100) and Key Projects of National Natural Science Foundation of China (U1836222 and 61733011).}  
	\\
	$^1$Department of Computer Science and Engineering, Shanghai Jiao Tong University\\
	$^2$Key Laboratory of Shanghai Education Commission for Intelligent Interaction\\
	and Cognitive Engineering, Shanghai Jiao Tong University, Shanghai, China\\
	{\tt\{zhangzs, huangyafang\}@sjtu.edu.cn},
	{\tt zhaohai@cs.sjtu.edu.cn}
}

\date{}

\begin{document}
\maketitle
\begin{abstract}
	Pinyin-to-character (P2C) conversion is the core component of pinyin-based Chinese input method engine (IME). However, the conversion is seriously compromised by the ambiguities of Chinese characters corresponding to pinyin as well as the predefined fixed vocabularies. To alleviate such inconveniences, we propose a neural P2C conversion model augmented by an online updated vocabulary with a sampling mechanism to support open vocabulary learning during IME working. Our experiments show that the proposed method outperforms commercial IMEs and state-of-the-art traditional models on standard corpus and true inputting history dataset in terms of multiple metrics and thus the online updated vocabulary indeed helps our IME effectively follows user inputting behavior.
\end{abstract}

\section{Introduction}

Chinese may use different Chinese characters up to 20,000 so that it is non-trivial to type the Chinese character directly from a Latin-style keyboard which only has 26 keys \cite{zhang2018SubMRC}. The pinyin as the official romanization representation for Chinese provides a solution that maps Chinese character to a string of Latin alphabets so that each character has a letter writing form of its own and users can type pinyin in terms of Latin letters to input Chinese characters into a computer. Therefore, converting pinyin to Chinese characters is the most basic module of all pinyin-based IMEs.

As each Chinese character may be mapped to a pinyin syllable, it is natural to regard the Pinyin-to-Character (P2C) conversion as a machine translation between two different languages, pinyin sequences and Chinese character sequences (namely Chinese sentence). Actually, such a translation in P2C procedure is even more straightforward and simple by considering that the target Chinese character sequence keeps the same order as the source pinyin sequence, which means that we can decode the target sentence from left to right without any reordering.

Meanwhile, there exists a well-known challenge in P2C procedure, too much ambiguity mapping pinyin syllable to character. In fact, there are only about 500 pinyin syllables corresponding to ten thousands of Chinese characters, even though the amount of the commonest characters is more than 6,000 \cite{Jia:2014}. As well known, the homophone and the polyphone are quite common in the Chinese language. Thus one pinyin may correspond to ten or more Chinese characters on the average.

However, pinyin IME may benefit from decoding longer pinyin sequence for more efficient inputting. When a given pinyin sequence becomes longer, the list of the corresponding legal character sequences will significantly reduce. For example, IME being aware of that pinyin sequence \emph{bei$\_$jing} can be only converted to either \begin{CJK*}{UTF8}{gkai} 背景 (\emph{background}) or 北京 (\emph{Beijing}) \end{CJK*} will greatly help it make the right and more efficient P2C decoding, as both pinyin \emph{bei} and \emph{jing} are respectively mapped to dozens of difference single Chinese characters.  Table \ref{tab:example} illustrates that the list size of the corresponding Chinese character sequence converted by pinyin sequence \begin{CJK*}{UTF8}{gkai}\emph{bei jing huan ying ni} (北 京 欢 迎 你, Welcome to Beijing) \end{CJK*} is changed according to the different sized source pinyin sequences.

\begin{CJK*}{UTF8}{gbsn}
	\begin{table}[!h]
			\centering
				\resizebox{\linewidth}{!}
		{
			\begin{tabular}{lccccc}
				\hline
				Pinyin seq. con-& bei & jing & huan & ying &ni \\
				sists of \textbf{1} syllable& 被 & 敬 & 环 & 英 & \textbf{你}\\
				& \textbf{北} & 静 & 换 & 颖 & 睨\\
				& 呗 & 井 & 还 & \textbf{迎} & 逆\\
				& 杯 & \textbf{京} & 幻 & 影 & 拟\\
				& 背 & 经 & \textbf{欢} & 应 & 尼\\
				\\
				Pinyin seq. con-&  \multicolumn{2}{l}{bei$\_$jing} & \multicolumn{2}{l}{huan$\_$ying} &ni \\
				sists of \textbf{2} syllables& \multicolumn{2}{l}{\textbf{北京}} & \multicolumn{2}{l}{幻影} & 你\\
				& \multicolumn{2}{l}{背景} & \multicolumn{2}{l}{\textbf{欢迎}} & 你\\
				\\
				Pinyin seq. con-&  \multicolumn{5}{l}{bei$\_$jing$\_$huan$\_$ying$\_$ni} \\
				sists of \textbf{5} syllables& \multicolumn{5}{l}{\textbf{北京欢迎你}}\\
				\hline
			\end{tabular}
		}
		\caption{\label{tab:example} The shorter the pinyin sequence is, the more character sequences will be mapped.}
	\end{table}
\end{CJK*}

To reduce the P2C ambiguities by decoding longer input pinyin sequence, Chinese IMEs may often utilize word-based language models since character-based language model always suffers from the mapping ambiguity. However, the effect of the work in P2C will be undermined with quite restricted vocabularies. The efficiency of IME conversion depends on the sufficiency of the vocabulary and previous work on machine translation has shown a large enough vocabulary is necessary to achieve good accuracy \cite{jean:2015}. In addition, some sampling techniques for vocabulary selection are proposed to balance the computational cost of conversion \cite{zhou:2016,wu:2018}. As IMEs work, users inputting style may change from time to time, let alone diverse user may input quite diverse contents, which makes a predefined fixed vocabulary can never be sufficient. For a convenient solution, most commercial IMEs have to manually update their vocabulary on schedule. Moreover, the training for word-based language model is especially difficult for rare words, which appear sparsely in the corpus but generally take up a large share of the dictionary.

To well handle the open vocabulary learning problem in IME, in this work, we introduce an online sequence-to-sequence (seq2seq) model for P2C and design a sampling mechanism utilizing our online updated vocabulary to enhance the conversion accuracy of IMEs as well as speed up the decoding procedure. In detail, first, a character-enhanced word embedding (CWE) mechanism is proposed to represent the word so that the proposed model can let IME generally work at the word level and pick a very small target vocabulary for each sentence. Second, every time the user makes a selection contradicted the prediction given by the P2C conversion module, the module will update the vocabulary accordingly. Our evaluation will be performed on three diverse corpora, including two which are from the real user inputting history, for verifying the effectiveness of the proposed method in different scenarios.

The rest of the paper is organized as follows: Section 2 discusses relevant works. Sections 3 and 4 introduce the proposed model. Experimental results and the model analysis are respectively in Sections 5 and 6. Section 7 concludes this paper.

\section{Related Work}


To effectively utilize words for IMEs, many natural language processing (NLP) techniques have been applied. \citet{chen:2003} introduced a joint maximum $n$-gram model with syllabification for grapheme-to-phoneme conversion. \citet{Chen2000A} used a trigram language model and incorporated word segmentation to convert pinyin sequence to Chinese word sequence. \citet{Xiaoqiang1996A} proposed an iterative algorithm to discover unseen words in corpus for building a Chinese language model. \citet{Mori2006Phoneme} described a method enlarging the vocabulary which can capture the context information.

For either pinyin-to-character for Chinese IMEs or kana-to-kanji for Japanese IMEs, a few language model training methods have been developed. \citet{Mori1998Kana} proposed a probabilistic based language model for IME. \citet{Jiampojamarn2008Joint} presented online discriminative training. \citet{Lin2008A} proposed a statistic model using the frequent nearby set of the target word. \citet{Chen2012Using} used collocations and k-means clustering to improve the n-pos model for Japanese IME. \citet{Jiang2007Pinyin} put forward a PTC framework based on support vector machine. \citet{Hatori2011Japanese} and \citet{Yang2012A} respectively applied statistic machine translation (SMT) to Japanese pronunciation prediction and Chinese P2C tasks. \citet{Shenyuan:2015,huang:2018acl}  regarded the P2C as a translation between two languages and solved it in neural machine translation framework.

All the above-mentioned work, however, still rely on a predefined fixed vocabulary, and IME users have no chance to refine their own dictionary through a user-friendly way. \citet{Zhang2017Tracing} is mostly related to this work, which also offers an online mechanism to adaptively update user vocabulary. The key difference between their work and ours lies on that this work presents the first neural solution with online vocabulary adaptation while \cite{Zhang2017Tracing} sticks to a traditional model for IME.

Recently, neural networks have been adopted for a wide range of tasks \cite{li2019dependency,xiao2019latt,zhou-2019-HPSG,li-etal-2018-seq2seq,li2018unified}. The effectiveness of neural models depends on the size of the vocabulary on the target side and previous work has shown that vocabularies of well over 50K word types are necessary to achieve good accuracy \cite{jean:2015} \cite{zhou:2016}. Neural machine translation (NMT) systems compute the probability of the next target word given both the previously generated target words as well as the source sentence. Estimating this conditional distribution is linear in the size of the target vocabulary which can be very large for many translation tasks. Recent NMT work has adopted vocabulary selection techniques from language modeling which do not directly generate the vocabulary from all the source sentences \cite{L2016Vocabulary,wu:2018}.

The latest studies on deep neural network prove the demonstrable effects of word representation on various NLP tasks, such as language modeling \cite{Verwimp2017Character}, question answering \cite{zhang2018OneShot,zhang2018char}, dialogue systems \cite{zhang2018DUA,Zhu2018lingke} and machine translation \cite{wang-etal-2017-sentence,wang-etal-2017-instance,wang-etal-2018-dynamic,8360031,AAAI1816060}. As for improved word representation in IMEs, \citet{Hatori2011Japanese} solved Japanese pronunciation inference combining word-based and character-based features within SMT-style framework to handle unknown words. \citet{Neubig2013Machine} proposed character-based SMT to handle sparsity. \citet{Okuno2012Using} introduced an ensemble model of word-based and character-based models for Japanese and Chinese IMEs. All the above-mentioned methods used similar solution about character representation for various tasks.

Our work takes inspiration from \cite{Luong2016Achieving} and \cite{Cai2017Fast}. The former built a novel representation method to tackle the rare word for machine translation. In detail, they used word representation network with characters as the basic input units. \citet{Cai2017Fast} presented a greedy neural word segmenter with balanced word and character embedding inputs. In the meantime, high-frequency word embeddings are attached to character embedding via average pooling while low-frequency words are computed from character embedding. Our embeddings also contain different granularity levels of embedding, but the word vocabulary is capable of being updated in accordance with users' inputting choice during IME working. In contrast, \cite{Cai2017Fast} build embeddings based on the word frequency from a fixed corpus.

\section{Our Models}

\begin{figure*}[!h]
	\centering
	\includegraphics[width=0.9\textwidth]{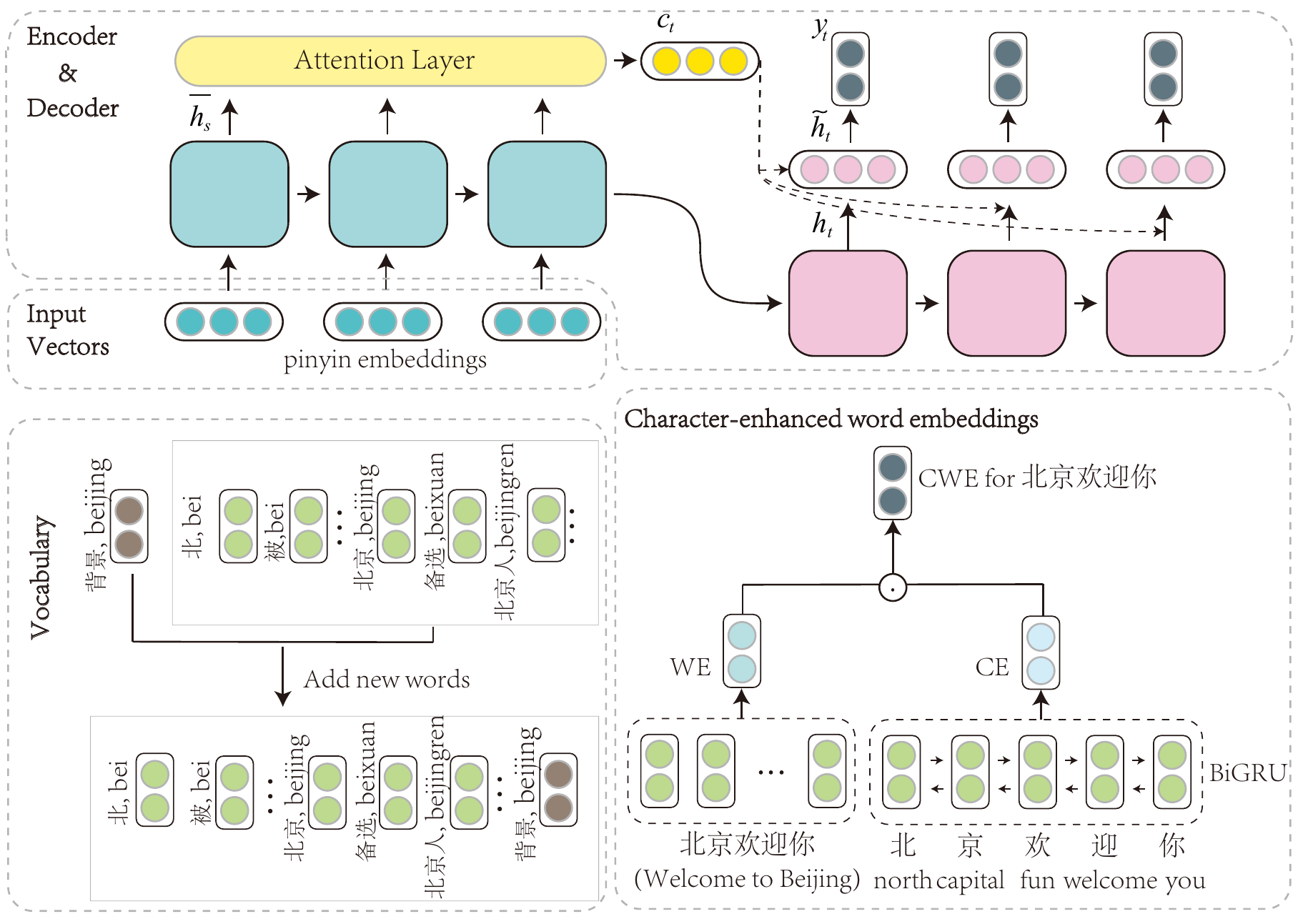}
	\caption{Architecture of the proposed Neural-based Chinese Input Method}
	\label{fig:framework}
\end{figure*}

For a convenient reference, hereafter a \emph{character} in pinyin language also refers to an independent pinyin syllable in the case without causing confusion, and \emph{word} means a pinyin syllable sequence which may correspond to a true word written in Chinese characters.

\begin{figure}[!h]
	\centering
	\includegraphics[width=0.3\textwidth]{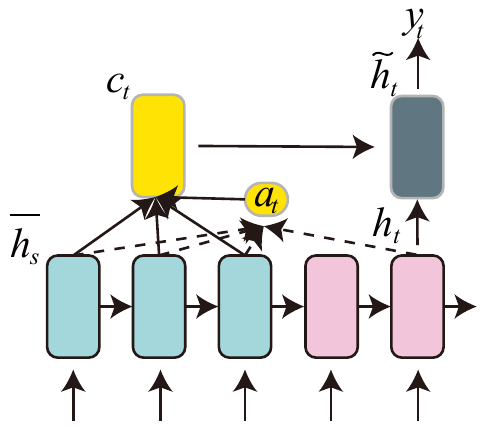}
	\caption{Architecture of the attention-based encoder-decoder model.}
	\label{fig:attlstm}
\end{figure}

As illustrated in Figure \ref{fig:framework}, the core of our hybrid P2C is a seq2seq model \cite{Cho2014Learning} in terms of the encoder-decoder framework. Given a pinyin sequence $X$ and a Chinese character sequence $Y$, the encoder of our neural P2C model utilizes a network for pinyin representation in which both word-level and character-level embedding are exploited, and the decoder is to generate the Chinese target sequence which maximizes $P(Y|X)$ using maximum likelihood training. 

Starting from an initial vocabulary with indicator from each turn of the user inputting choice, the online learning module helps update the word vocabulary by minimizing empirical prediction risk.


\subsection{Pinyin-Character Parallel Corpus}

Pinyin-character parallel corpus can be conveniently generated by automatically annotating Chinese character text with pinyin as introduced in \cite{Yang2012A}. Using standard Chinese word segmentation methods, we may segment both character and pinyin text into words with the same segmentation for each sentence pair.

\subsection{Encoder-Decoder}

The encoder is a bi-directional long short-term memory (LSTM) network (Hochreiter and Schmidhuber, 1997). The vectorized inputs are fed to forward and backward LSTMs to obtain the internal representation of two directions. The output for each input is the concatenation of the two vectors from both directions. Our decoder is based on the global attentional models proposed by \citet{Luong2015Effective} to consider the hidden states of the encoder when deriving the context vector. The probability is conditioned on a distinct context vector for each target word. The context vector is computed as a weighted sum of previous hidden states. The probability of each candidate word as being the recommended one is predicted using a softmax layer over the inner-product between source embeddings and candidate target characters. Figure \ref{fig:attlstm} shows the architecture.

\section{Online P2C Learning with Vocabulary Adaptation}\label{sec:cew}

\begin{figure}[!h]
	\centering
	\includegraphics[width=0.25\textwidth]{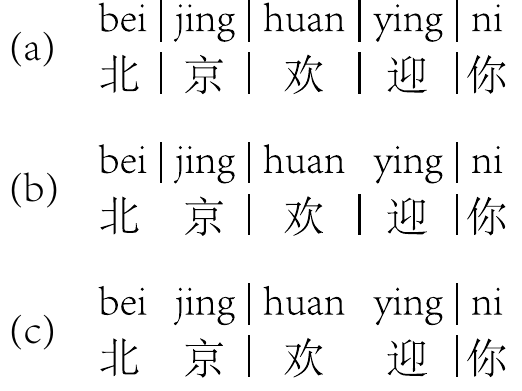}
	\caption{Different segmentations decide different alignments}
	\label{fig:cewwc}
\end{figure}

As the core of Chinese IME, P2C conversion has been formulized into a seq2seq model as machine translation between pinyin and character sequences, there are still a few differences between P2C converting and standard machine translation. 1) Considering both pinyin syllables and Chinese characters are segmented into single-character \emph{word} as Figure \ref{fig:cewwc}a, there is a one-to-one mapping between any character and its corresponding pinyin syllable without word reordering, while typical machine translation does not enjoy such benefits and has to perform careful word reordering explicitly or implicitly. 2) As Chinese language is always sensitive to the segmentation scheme, in the writing of either the Chinese character or the pinyin, P2C as NMT may suffer from alignment mismatch on both sides like Figure \ref{fig:cewwc}b or benefit a lot from perfect one-to-one alignment like Figure \ref{fig:cewwc}c, while typical machine translation is seldom affected by such segmentation alignment. 3) P2C as a working component of IME, every time it returns a list of Chinese character sequence predictions, user may indicate which one is what he or she actually expects to input. To speed up the inputting, IME always tries to rank the user's intention at the top-1 position. So does IME, we say there is a correct conversion or prediction. Different from machine translation job, users' inputting choice will always indicate the `correct' prediction right after IME returns the list of its P2C conversion results.

Therefore, IME working mode implies an online property, we will let our neural P2C model also work and evaluate in such a way. Meanwhile, online working means that our model has to track the continuous change of users' inputting contents, which is equally a task about finding new words in either pinyin sequence or character sequence.

However, the word should be obtained through the operation of word segmentation. Note that as we have discussed above, segmentation over pinyin sequence is also necessary to alleviate the ambiguity of pinyin-to-character mapping. Thus the task here for IME online working actually requires an online word segmentation algorithm if we want to keep the aim of open vocabulary learning. Our solution to this requirement is adopting a vocabulary-based segmentation approach, namely, the maximum matching algorithm which greedily segments the longest matching in the given vocabulary at the current segmentation point of a given sequence. Then adaptivity of the segmentation thus actually relies on the vocabulary online updating.

Algorithm \ref{alg:adaoalgo} gives our online vocabulary updating algorithm for one-turn IME inputting. Note the algorithm maintains a pinyin-character bilingual vocabulary. Collecting the user's inputting choices through IME, our P2C model will perform online training over segmented pinyin and character sequences with the updated vocabulary. The updating procedure introduces new words by comparing the user's choice and IME's top-1 prediction. The longest mismatch $n$-gram characters will be added as new word.

\begin{algorithm}[!htb]
	\caption{Online Vocabulary Updating Algorithm}
	\label{alg:adaoalgo}
	\begin{algorithmic}[1] 
		\REQUIRE ~~\\
		\begin{itemize}[leftmargin=*]
			\item Vocabulary: $V = \{(Py_i, Ch_i)|i=1,2,3,\cdots\}$;
			\item Input pinyin sequence: $Py=\{py_i|i=1,2,3,\cdots\}$;
			\item IME predicted top-1 character sequence: $Cm=\{cm_i|i=1,2,3,\cdots\}$;
			\item User choosing character sequence: $Cu=\{cu_i|i=1,2,3,\cdots\}$.
		\end{itemize}
		\ENSURE ~~\\ 
		\begin{itemize}[leftmargin=*]
			\item The Updated Vocabulary: $\hat{V}$.
		\end{itemize}
		\STATE $\rhd$ Adding new words
		\FOR{$n = 6$ to $2$}
		\STATE Compare $n$-gram of $Cu$ and $Cm$
		\IF{Mismatch $Ch$ is found // both the first and last characters are different at least}
		\IF{$Ch$ is not in $\hat{V}$}
		\STATE $V = V \cup \{Ch\}$
		\ENDIF
		\ENDIF
		\IF{no mismatch is found}
		\BREAK
		\ENDIF
		\ENDFOR
		\RETURN $\hat{V}$; 
	\end{algorithmic}
\end{algorithm}
\renewcommand{\baselinestretch}{1.0}

We adopt a hybrid mechanism to balance both words and characters representation, namely, Character-enhanced Word Embedding (CWE). In the beginning, we keep an initial vocabulary with the most frequent words. The words inside the vocabulary are represented as enhanced-embedding, and those outside the list are computed from character embeddings. A pre-trained word2vec model \cite{mikolov:2013} is generated to represent the word embedding $WE(w)$($w\in \hat{V}$). At the same time we feed all characters of each word to a bi-gated recurrent unit (bi-GRU) \cite{Cho2014Learning} to compose the character level representation $CE(w)$ ($w = \{c_i|i=1,2,3,\cdots\}$).

The enhanced embedding $CWE(w)$ is to straightforwardly integrate word embedding  and character embedding by element-wise multiplication, 
\begin{equation*}
CWE(w) = WE(w) \odot CE(w)
\end{equation*}

\section{Target Vocabulary Selection}\label{sec:tvs}

In this section, we aim to prune the target vocabulary $\hat{V}$ as small as possible to reduce the computing time. Our basic idea is to maintain a separate and small vocabulary $\hat{V}$ for each sentence so that we only need to compute the probability distribution over a small vocabulary for each sentence.

We first generate a sentence-level vocabulary $V_s$ to be one part of our $\hat{V}$, which includes the mapped Chinese words of each pinyin in the source sentence. As the bilingual vocabulary $V$ consists of the pinyin and Chinese word pair of all the words that ever appeared, it is natural to use a prefix maximum matching algorithm to obtain a sorted list of relevant candidate translations $D(x) = [Ch_1, Ch_2, ...]$ for the source pinyin. Thus, we generate a target vocabulary $V_s$ for a sentence $x = (Py_1, Py_2, ...)$ by merging all the candidates of all pinyin.

In order to cover target un-aligned functional words, we also need top $n$ most common target words $V_c$.

In training procedure, the target vocabulary $\hat{V}$ for a sentence $x$ needs to include the target words $V_t$ in the reference $y$, $\hat{V} = V_s \cup V_c \cup V_y$.

In decoding procedure, the $\hat{V}$ may only contain two parts,
$\hat{V} = V_s \cup V_c$.

\section{Experiment}

\subsection{Datasets and Evaluation Metrics}

We adopt two corpora for evaluation. The People's Daily corpus is extracted from the People's Daily from 1992 to 1998 by Peking University \cite{thomas2005the}. The bilingual corpus can be straightforwardly produced by the conversion proposed by \cite{Yang2012A}. Contrast to the style of the People's Daily, the TouchPal corpus \cite{Zhang2017Tracing} is a large scale of user chat history collected by TouchPal IME, which are more colloquial. Hence, we use the latter to simulate user's chatting input to verify our online model's adaptability to different environments. The test set size is 2,000 MIUs in both corpora. Table \ref{tab:dataset} shows the statistics of two corpora\footnote{The two corpora along with our codes are available at \url{https://github.com/cooelf/OpenIME} .}.

\begin{table}[!hb]
	\centering
	{
		\begin{tabular}{p{0.3cm}|l|c|c}
			\hline
			\hline
			& & Chinese & Pinyin \\ \cline{3-4}
			\hline
			\multirow{3}{*}{PD} & $\#$ MIUs & \multicolumn{2}{c}{5.04M} \\ \cline{3-4}
			& $\#$ Word & 24.7M & 24.7M \\ \cline{3-4}
			& $\#$ Vocab & 54.3K & 41.1K  \\ \cline{3-4}
			& $\#$ Target Vocab (train)& 2309 & - \\ \cline{3-4}
			& $\#$ Target Vocab (dec)& 2168 & - \\ \cline{3-4}
			\hline
			
			\multirow{3}{*}{TP} & $\#$ MIUs & \multicolumn{2}{c}{689.6K} \\ \cline{3-4}
			& $\#$ Word & 4.1M & 4.1M \\ \cline{3-4}
			& $\#$ Vocab & 27.7K & 20.2K \\ \cline{3-4}
			& $\#$ Target Vocab (train)& 2020 & - \\ \cline{3-4}
			& $\#$ Target Vocab (dec)& 2009 & - \\ \cline{3-4}
			\hline
			
			\hline
		\end{tabular}
	}
	\caption{\label{tab:dataset} MIUs count, word count and vocab size statistics of our training data. PD refers to the People's Daily, TP is TouchPal corpus.}
\end{table}
\renewcommand{\baselinestretch}{1.0}

Two metrics are used for our evaluation by following previous work: Maximum Input Unit (MIU) Accuracy and KeyStroke Score (KySS) \cite{Jia2013Kyss}. The former measures the conversion accuracy of MIU, which is defined as the longest uninterrupted Chinese character sequence inside a sentence. As the P2C conversion aims to output a rank list of corresponding character sequences candidates, the top-$K$ MIU accuracy means the possibility of hitting the target in the first $K$ predict items. We will follow the definition of \cite{Zhang2017Tracing} about top-$K$ accuracy. The KySS quantifies user experience by using keystroke count. An IME with higher KySS is supposed to perform better. For an ideal IME, there will be KySS$\ = 1$.

\begin{table*}[!h]
	\centering
	{
		\begin{tabular}{c|c|ccc|ccc}
			\hline
			\hline
			\multirow{2}{*}{System} & \multirow{2}{*}{ED} & \multicolumn{3}{c}{PD} & \multicolumn{3}{c}{TP}\\ \cline{3-8}
			&& Top1 & Top5 & Top10 & Top1 & Top5 & Top10 \\
			\hline
			\multicolumn{8}{l}{Existing P2C} \\
			\hline
			Google IME && 70.9 & 78.3 & \textbf{82.3} & 57.5 & 63.8 & 69.3\\
			OMWA && 55.0 & 63.7 & 70.2 & 19.7 & 24.8 & 27.7 \\
			On-OMWA&& 64.4 & 72.9 & 77.9 & 57.1 & 71.1 & 80.9\\
			\hline
			\multicolumn{8}{l}{Our P2C} \\
			\hline
			Base P2C&200& 53.2 & 64.7 & 70.3 & 46.8 & 68.8 & 75.7\\
			On-P2C&200& 68.1 & 77.3 & 78.2 & 69.8 & 88.7 & 89.3\\
			On-P2C (bi)&200& 70.5 & 79.8 & 80.1 & 71.0 & 89.2 & 89.5\\
			On-P2C (bi)&300& 70.8 & \textbf{80.5} & 81.2 & \textbf{71.9} & 89.6 & \textbf{90.6}\\
			On-P2C (bi)&400&\textbf{71.3} & 80.1 & \textbf{81.3} & 71.7 & \textbf{89.7} & 90.3\\
			On-P2C (bi)&500&69.9 & 78.2 & 81.0 & 70.7 & 89.2 & 89.8 \\
			\hline
			\hline
		\end{tabular}
	}
	\caption{\label{tab:result1}Top-$K$ accuracies on the People's Daily (PD) , TouchPal (TP) corpora. ED refers to embedding dimension. The best results are in bold.}
\end{table*}
\renewcommand{\baselinestretch}{1.0}
\subsection{Settings}

IME works giving a list of character sequence candidates for user choosing. Therefore, measuring IME performance is equivalent to evaluating such a rank list. In this task, we select 5 converted character sequence candidates for each pinyin sequence. Given a pinyin sequence and candidate characters, our model is designed to rank the characters in an appropriate order.

Here is the model setting we used: a) pre-trained word embeddings were generated on the People's Daily corpus; b) the recurrent neural networks for encoder and decoder have 3 layers and 500 cells, and the representation networks have 1 layer; c) the initial learning rate is 1.0, and we will halve the learning rate every epoch after 9 epochs; d) dropout is 0.3; e) the default frequency filter ratio for CWE establishment is 0.9. The same setting is applied to all models.

For a balanced treatment over both corpora, we used baseSeg \cite{Hai2006An} to segment all text, then extract all resulted words into the initial vocabulary for online evaluation. We train the base P2C models for 13 epochs with plain stochastic gradient descent on the People's Daily corpus with 32 batch size, and the online training process runs 25 epochs with 1 batch size. In practical application, we perform online training for once every 64 instances are inputs to control costs.

\begin{figure*}[!h]
	\centering
	\includegraphics[width=1.0\textwidth]{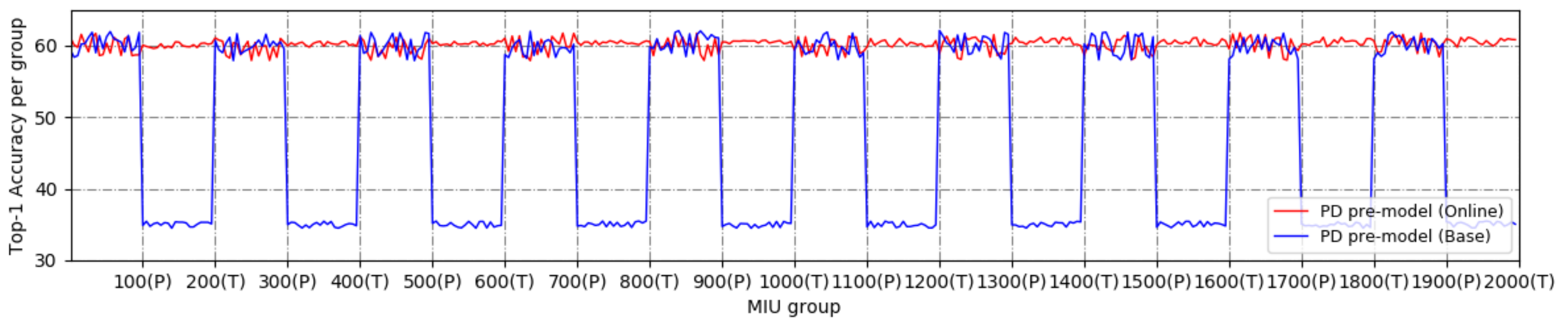}
	\caption{Top-1 accuracy on an interlaced joint Corpus. P: the People's daily segment, T: Touchpal segment.}
	\label{fig:result6}
\end{figure*}

\subsection{Results}

\begin{table*}[!h]
	\centering
	{
		\begin{tabular}{l|ccccc}
			\hline
			\hline
			Filter Ratio & 0 & 0.3 & 0.6 & 0.9 & 1.0 \\
			Top-5 Accuracy(valid set) & 66.4 & 68.3 & 84.3 & 89.7 & 87.5 \\
			Top-5 Accuracy(test set) & 66.3 & 68.1 & 83.9 & \textbf{89.6} & 87.1 \\
			\hline
			\hline
		\end{tabular}
	}
	\caption{\label{tab:result3} Top-5 accuracies of P2C after filtering specific ratio of words from vocabulary.}
\end{table*}

\begin{table}[!h]
	\centering
	{
		\begin{tabular}{l|c|c}
			\hline
			\hline
			Models & the People's Daily & TouchPal\\
			\hline
			Google IME & 0.7535 & 0.6465 \\
			OMWA & 0.6496 & 0.4489 \\
			On-OMWA & 0.7115 & 0.7226 \\
			\hline
			Base P2C & 0.6922 & 0.7910 \\
			On-P2C & \textbf{0.8301} & \textbf{0.8962} \\
			\hline
			\hline
		\end{tabular}
	}
	\caption{\label{tab:kyssresult} User experience in terms of KySS}
\end{table}
\renewcommand{\baselinestretch}{1.0}

We compare our P2C conversion system with two baseline systems, Google IME \footnote{The Google IME is the only commercial Chinese IME providing a debuggable API on the market now.} and Offline and Online models for Word Acquisition (OMWA, On-OMWA)\cite{Zhang2017Tracing}, and the results are shown in Table \ref{tab:result1}.

On the People's Daily corpus, our online model (On-P2C) outperforms the best model in \cite{Zhang2017Tracing} by +3.72$\%$ top-1 MIU accuracy. The +14.94 improvement over the base P2C conversion module demonstrates that online learning vocabulary is effective. The using of bi-direction LSTM encoder produces a notable boost of +2.41$\%$ accuracy. Our P2C model seizes a slight but significant improvement when tuning the dimension of CWE; our model gives 71.32$\%$ top-1 MIU accuracy. The performance on TouchPal corpus is similar and even more obvious; our best setting achieves 14.35$\%$ improvements compared to the best baseline.

The P2C module of IME outputs a rank list, and then the IME once displays five candidates by default. If users cannot find the target character in the top 5 candidates, they have to click the Page Down button to navigate more candidates, which involve additional keystroke expenses for users. Therefore, we list the top-5 accuracy contrast to all baselines with top-10 results, and the comparison indicates the noticeable advancement of our P2C model. On TouchPal corpus, our model with the best setting achieves 89.7$\%$ accuracy, surpassing all the baselines.

\section{Analysis}

\subsection{Effects of Online Updated Vocaburay}
\begin{figure}[!h]
	\centering
	\includegraphics[width=0.5\textwidth]{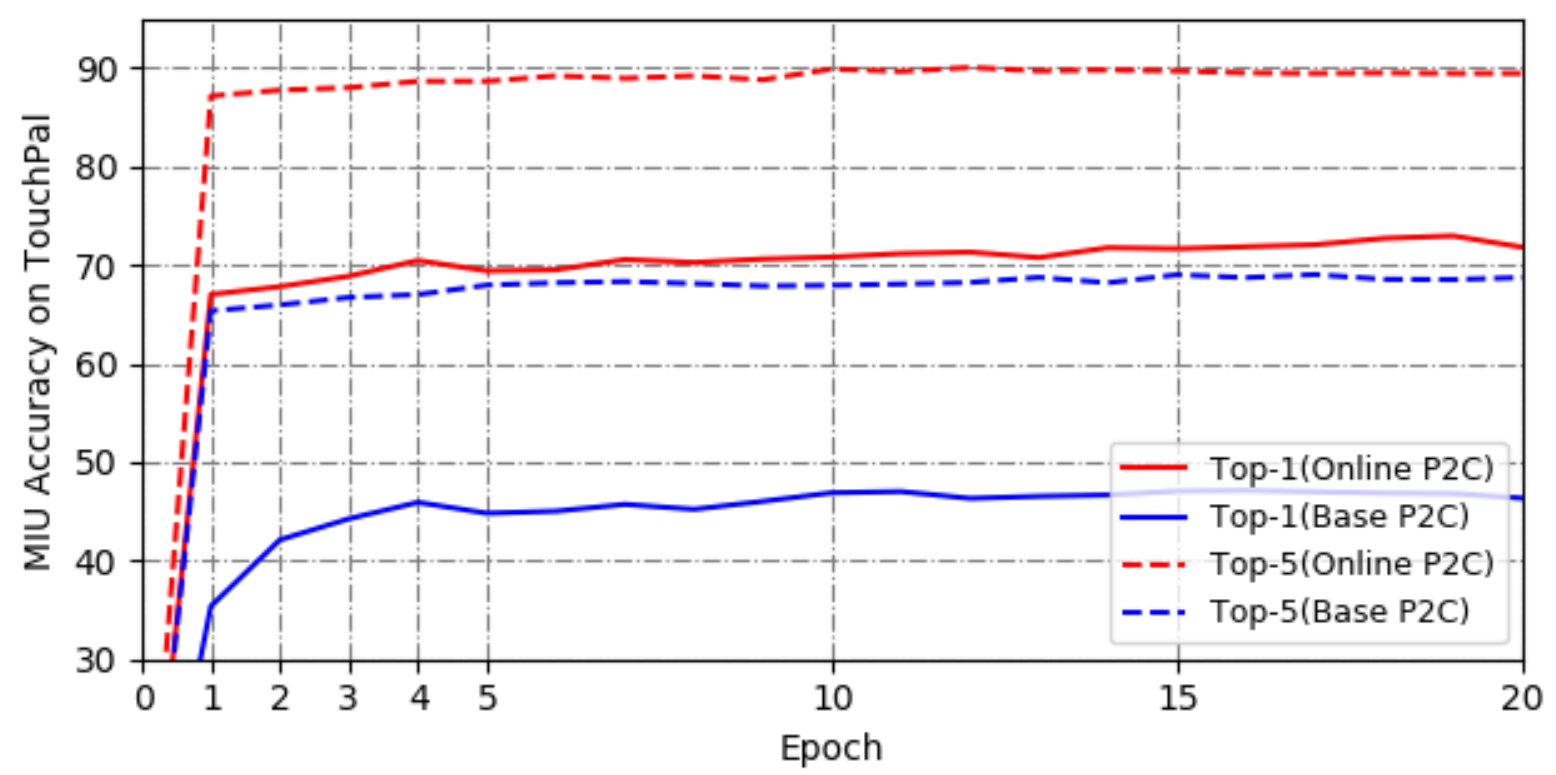}
	\caption{Training curves of top-1 and top-5 accuracy on TouchPal.}
	\label{fig:result2}
\end{figure}

Figure \ref{fig:result2} shows the changes of the MIU accuracy during the training process. For both top-1 and top-5 MIU accuracy, models with online vocabulary updating significantly outperform those without updating throughout the entire training. Especially, online P2C gives top-1 MIU accuracy comparable to top-5 MIU accuracy given by the base P2C module, which suggests a great inputting efficiency improvement from introducing the online updating mechanism.

Figure \ref{fig:result6} expounds the adaptivity of our online P2C, in which we feed a joint corpus that is extracted from test corpora of the People's Daily and Touchpal to the base P2C model and record the top-1 MIU accuracy per group after 2 epochs online vocabulary learning with batch size 1. We see that online P2C distinctly adapts the corpus change at the joint part. On the contrary, the base P2C which works offline performs stably only on its in-domain segments.

\subsection{Effects of Vocaburay Selection}

\begin{figure}[!h]
	\centering
	\includegraphics[width=0.5\textwidth]{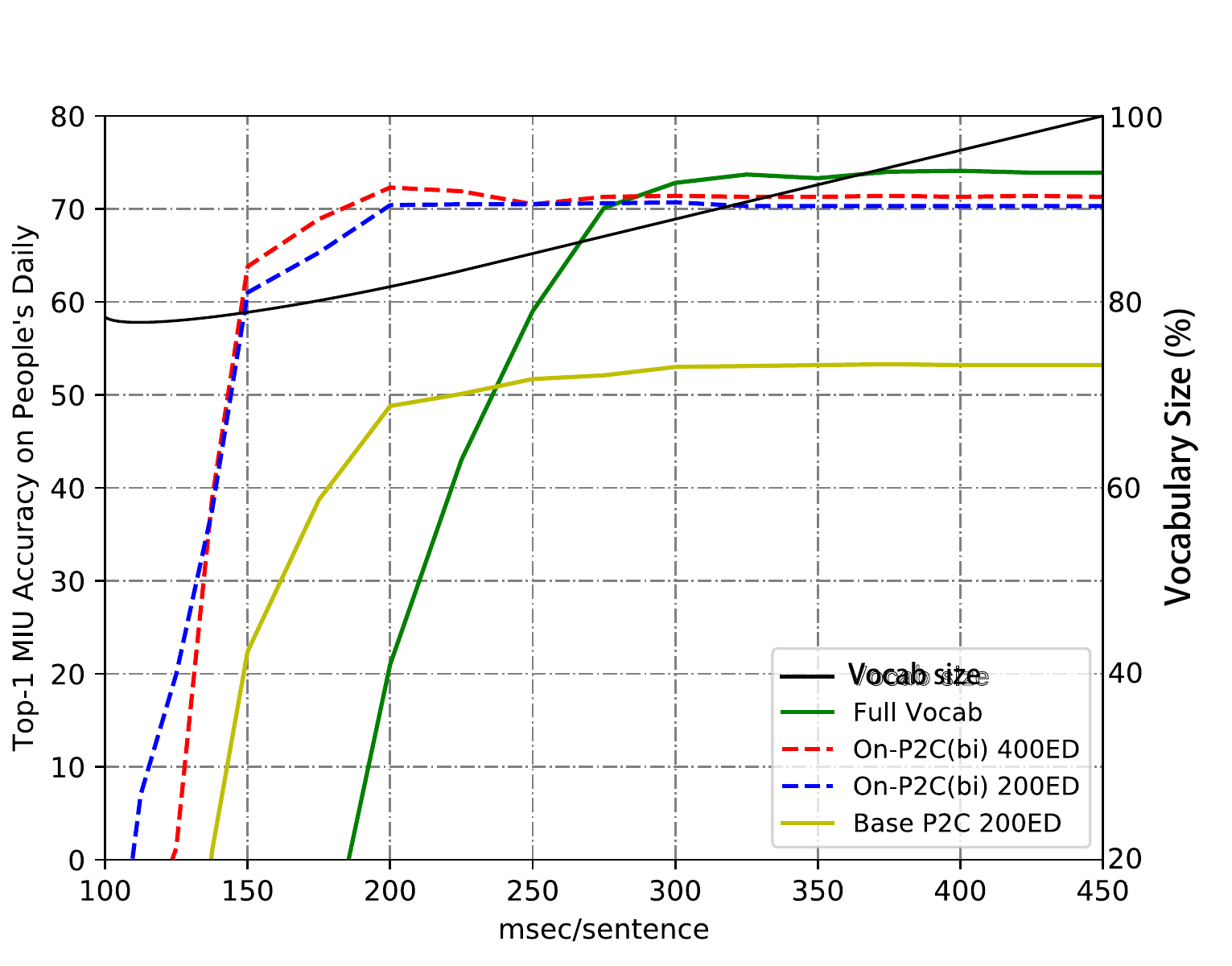}
	\caption{MIU accuracy versus decoding time on CPU.}
	\label{fig:result5}
\end{figure}

As an adaptive vocabulary is used in our decoder, it may result in a very large vocabulary to encumber the decoder with efficiency. Therefore, in practice, we need to control the size of the vocabulary for acceptable decoding speed. However, pruning the vocabulary in any way will surely hurt the performance due to all items in the adaptive vocabulary added with a reason. Figure  \ref{fig:result5}  illustrates the relation between accuracy and decoding speed. The accuracies nearly do not get decreased with high enough decoding speed when only taking 88.9\% full vocabulary in our system.

\subsection{Effects of Word Filtering for CWE building}

As we mentioned in Section \ref{sec:cew}, P2C conversion quality depends on the CWE mechanism which will benefit from an appropriate filtration ratio. As shown in Table \ref{tab:result3}, when the filter ratio equals to 0.9, the accuracy reaches the top. We notice two observations. First, pure word-level representation is more efficient for P2C tasks than character-level which only achieves 66.3$\%$ accuracy. Second, omitting partial low-frequency word is instrumental in establishing word-level embedding. Actually, when building word embeddings, rare words behave no more than noise. If the rare words are not initialized properly, they would also bias the whole word embeddings. Therefore, we more incline to make character-level embedding to represent a rare word, and build CWE embeddings for others.

\subsection{User Experience}
\citet{Jia2013Kyss} proposed that the user-IME interaction contains three steps: pinyin input, candidate index choice and page turning. In Table \ref{tab:result1}, the $89.7\%$ top-5 accuracy on TouchPal means that users have nearly 90$\%$ possibilities to straightly obtain the expected inputs in the first page (usually 5 candidates per page for most IME interface setting), so that user experiment using IMEs can be directly measured by KySS. Table \ref{tab:kyssresult} shows the mean KySS of various models. The results indicate that our P2C conversion module further facilitates the interaction.

\section{Conclusion}

This paper presents the first neural P2C converter for pinyin-based Chinese IME with open vocabulary learning as to our best knowledge. We adopt an online working-style seq2seq model for the concerned task by formulizing it as a machine translation from pinyin sequence to Chinese character sequence. In addition, we propose an online vocabulary updating algorithm for further performance enhancement by tracking users behavior effectively. The evaluation on the standard linguistic corpus and true inputting history show the proposed methods indeed greatly improve user experience in terms of diverse metrics compared to commercial IME and state-of-the-art traditional model.
\bibliography{acl2019}

\begin{thebibliography}{45}
\expandafter\ifx\csname natexlab\endcsname\relax\def\natexlab#1{#1}\fi

\bibitem[{Cai et~al.(2017)Cai, Zhao, Zhang, Xin, Wu, and Huang}]{Cai2017Fast}
Deng Cai, Hai Zhao, Zhisong Zhang, Yuan Xin, Yongjian Wu, and Feiyue Huang.
  2017.
\newblock \href {https://doi.org/10.18653/v1/P17-2096} {Fast and accurate
  neural word segmentation for {Chinese}}.
\newblock In \emph{Proceedings of the 55th Annual Meeting of the Association
  for Computational Linguistics (ACL)}, pages 608--615.

\bibitem[{Chen et~al.(2018)Chen, Wang, Utiyama, Sumita, and Zhao}]{AAAI1816060}
Kehai Chen, Rui Wang, Masao Utiyama, Eiichiro Sumita, and Tiejun Zhao. 2018.
\newblock \href
  {https://www.aaai.org/ocs/index.php/AAAI/AAAI18/paper/view/16060/16008}
  {Syntax-directed attention for neural machine translation}.
\newblock In \emph{Proceedings of the AAAI Conference on Artificial
  Intelligence (AAAI)}, pages 4792--4799.

\bibitem[{Chen et~al.(2012)Chen, Wu, and He}]{Chen2012Using}
Long Chen, Xianchao Wu, and Jingzhou He. 2012.
\newblock \href {https://www.aclweb.org/anthology/W12-4804} {Using collocations
  and k-means clustering to improve the n-pos model for japanese ime}.
\newblock In \emph{Proceedings of the Second Workshop on Advances in Text Input
  Methods}, pages 45--56.

\bibitem[{Chen et~al.(2015)Chen, Wang, and Zhao}]{Shenyuan:2015}
Shenyuan Chen, Rui Wang, and Hai Zhao. 2015.
\newblock \href {https://www.aclweb.org/anthology/Y15-1052} {Neural network
  language model for {Chinese} pinyin input method engine}.
\newblock In \emph{Proceedings of the 29th Pacific Asia Conference on Language,
  Information and Computation (PACLIC)}, pages 455--461.

\bibitem[{Chen(2003)}]{chen:2003}
Stanley~F. Chen. 2003.
\newblock \href {https://doi.org/10.1109/ICASSP.2017.7952674} {Conditional and
  joint models for grapheme-to-phoneme conversion}.
\newblock In \emph{Eighth European Conference on Speech Communication and
  Technology}, pages 2033--2036.

\bibitem[{Chen and Lee(2000)}]{Chen2000A}
Zheng Chen and Kai~Fu Lee. 2000.
\newblock \href {https://doi.org/10.3115/1075218.1075249} {A new statistical
  approach to {Chinese} pinyin input}.
\newblock In \emph{Proceedings of the 38th Annual Meeting of the Association
  for Computational Linguistics (COLING)}, pages 241--247.

\bibitem[{Cho et~al.(2014)Cho, Merrienboer, Gulcehre, Bahdanau, Bougares,
  Schwenk, and Bengio}]{Cho2014Learning}
Kyunghyun Cho, Bart~Van Merrienboer, Caglar Gulcehre, Dzmitry Bahdanau, Fethi
  Bougares, Holger Schwenk, and Yoshua Bengio. 2014.
\newblock \href {https://www.aclweb.org/anthology/D14-1179} {Learning phrase
  representations using {RNN} encoder-decoder for statistical machine
  translation}.
\newblock In \emph{Proceedings of the 2014 Conference on Empirical Methods in
  Natural Language Processing (EMNLP)}, pages 1724--1734.

\bibitem[{Emerson(2005)}]{thomas2005the}
Thomas Emerson. 2005.
\newblock \href {https://aclweb.org/anthology/I05-3017} {The second
  international {Chinese} word segmentation bakeoff}.
\newblock In \emph{Proceedings of the fourth SIGHAN workshop on Chinese
  language Processing}, pages 123--133.

\bibitem[{Hatori and Suzuki(2011)}]{Hatori2011Japanese}
Jun Hatori and Hisami Suzuki. 2011.
\newblock \href {https://www.aclweb.org/anthology/I11-1014} {{Japanese}
  pronunciation prediction as phrasal statistical machine translation}.
\newblock In \emph{Proceedings of 5th International Joint Conference on Natural
  Language Processing (IJCNLP)}, pages 993--1004.

\bibitem[{Huang et~al.(2018)Huang, Li, Zhang, and Zhao}]{huang:2018acl}
Yafang Huang, Zuchao Li, Zhuosheng Zhang, and Hai Zhao. 2018.
\newblock \href {https://www.aclweb.org/anthology/P18-4024} {{Moon IME:}
  neural-based chinese pinyin aided input method with customizable
  association}.
\newblock In \emph{Proceedings of the 56th Annual Meeting of the Association
  for Computational Linguistics (ACL), System Demonstration}, pages 140--145.

\bibitem[{Jean et~al.(2015)Jean, Cho, Memisevic, and Bengio}]{jean:2015}
Sebastien Jean, Kyunghyun Cho, Roland Memisevic, and Yoshua Bengio. 2015.
\newblock \href {https://doi.org/10.3115/v1/P15-1001} {On using very large
  target vocabulary for neural machine translation}.
\newblock In \emph{Proceedings of the 53rd Annual Meeting of the Association
  for Computational Linguistics and the 7th International Joint Conference on
  Natural Language Processing (ACL-IJCNLP)}, pages 1--10.

\bibitem[{Jia and Zhao(2013)}]{Jia2013Kyss}
Zhongye Jia and Hai Zhao. 2013.
\newblock \href {https://www.aclweb.org/anthology/I13-1170} {Kyss 1.0: a
  framework for automatic evaluation of {Chinese} input method engines}.
\newblock In \emph{Proceedings of the Sixth International Joint Conference on
  Natural Language Processing (IJCNLP)}, pages 1195--1201.

\bibitem[{Jia and Zhao(2014)}]{Jia:2014}
Zhongye Jia and Hai Zhao. 2014.
\newblock \href {https://doi.org/10.3115/v1/P14-1142} {A joint graph model for
  pinyin-to-chinese conversion with typo correction}.
\newblock In \emph{Proceedings of the 52nd Annual Meeting of the Association
  for Computational Linguistics (ACL)}, pages 1512--1523.

\bibitem[{Jiampojamarn et~al.(2008)Jiampojamarn, Cherry, and
  Kondrak}]{Jiampojamarn2008Joint}
Sittichai Jiampojamarn, Colin Cherry, and Grzegorz Kondrak. 2008.
\newblock \href {https://www.aclweb.org/anthology/P08-1103} {Joint processing
  and discriminative training for letter-to-phoneme conversion}.
\newblock In \emph{Proceedings of ACL-08: HLT}, pages 905--913.

\bibitem[{Jiang et~al.(2007)Jiang, Guan, Wang, and Liu}]{Jiang2007Pinyin}
Wei Jiang, Yi~Guan, Xiao~Long Wang, and Bing~Quan Liu. 2007.
\newblock \href {https://doi.org/10.1007/s10822-007-9102-6} {Pinyin to
  character conversion model based on support vector machines}.
\newblock \emph{Journal of Chinese Information Processing}, 21(2):100--105.

\bibitem[{L'Hostis et~al.(2016)L'Hostis, Grangier, and Auli}]{L2016Vocabulary}
Gurvan L'Hostis, David Grangier, and Michael Auli. 2016.
\newblock \href {https://arxiv.org/pdf/1610.00072.pdf} {Vocabulary selection
  strategies for neural machine translation}.
\newblock \emph{arXiv preprint arXiv:1610.00072}.

\bibitem[{Li et~al.(2018{\natexlab{a}})Li, Cai, He, and
  Zhao}]{li-etal-2018-seq2seq}
Zuchao Li, Jiaxun Cai, Shexia He, and Hai Zhao. 2018{\natexlab{a}}.
\newblock \href {https://www.aclweb.org/anthology/C18-1271} {Seq2seq dependency
  parsing}.
\newblock In \emph{Proceedings of the 27th International Conference on
  Computational Linguistics (COLING)}, pages 3203--3214.

\bibitem[{Li et~al.(2018{\natexlab{b}})Li, He, Cai, Zhang, Zhao, Liu, Li, and
  Si}]{li2018unified}
Zuchao Li, Shexia He, Jiaxun Cai, Zhuosheng Zhang, Hai Zhao, Gongshen Liu,
  Linlin Li, and Luo Si. 2018{\natexlab{b}}.
\newblock \href {https://aclweb.org/anthology/D18-1262} {A unified syntax-aware
  framework for semantic role labeling}.
\newblock In \emph{Proceedings of the 2018 Conference on Empirical Methods in
  Natural Language Processing (EMNLP)}, pages 2401--2411.

\bibitem[{Li et~al.(2019)Li, He, Zhao, Zhang, Zhang, Zhou, and
  Zhou}]{li2019dependency}
Zuchao Li, Shexia He, Hai Zhao, Yiqing Zhang, Zhuosheng Zhang, Xi~Zhou, and
  Xiang Zhou. 2019.
\newblock \href {https://arxiv.org/pdf/1901.05280v1.pdf} {Dependency or span,
  end-to-end uniform semantic role labeling}.
\newblock \emph{arXiv preprint arXiv:1901.05280}.

\bibitem[{Lin and Zhang(2008)}]{Lin2008A}
Bo~Lin and Jun Zhang. 2008.
\newblock \href {https://doi.org/10.1145/1458082.1458318} {A novel statistical
  {Chinese} language model and its application in pinyin-to-character
  conversion}.
\newblock In \emph{Proceedings of the 17th ACM conference on Information and
  knowledge management}, pages 1433--1434.

\bibitem[{Luong and Manning(2016)}]{Luong2016Achieving}
Minh-Thang Luong and Christopher~D. Manning. 2016.
\newblock \href {https://doi.org/10.18653/v1/P16-1100} {Achieving open
  vocabulary neural machine translation with hybrid word-character models}.
\newblock In \emph{Proceedings of the 54th Annual Meeting of the Association
  for Computational Linguistics (ACL)}, pages 1054--1063.

\bibitem[{Luong et~al.(2015)Luong, Pham, and Manning}]{Luong2015Effective}
Minh~Thang Luong, Hieu Pham, and Christopher~D Manning. 2015.
\newblock \href {https://doi.org/10.18653/v1/D15-1166} {Effective approaches to
  attention-based neural machine translation}.
\newblock In \emph{Proceedings of the 2015 Conference on Empirical Methods in
  Natural Language Processing (EMNLP)}, pages 1412--1421.

\bibitem[{Mikolov et~al.(2013)Mikolov, Chen, Corrado, and Dean}]{mikolov:2013}
Tomas Mikolov, Kai Chen, Greg Corrado, and Jeffrey Dean. 2013.
\newblock \href {https://arxiv.org/pdf/1301.3781.pdf} {Efficient estimation of
  word representations in vector space}.
\newblock \emph{arXiv preprint arXiv:1301.3781}.

\bibitem[{Mori et~al.(2006)Mori, Takuma, and Kurata}]{Mori2006Phoneme}
Shinsuke Mori, Daisuke Takuma, and Gakuto Kurata. 2006.
\newblock \href {https://doi.org/10.3115/1220175.1220267} {Phoneme-to-text
  transcription system with an infinite vocabulary}.
\newblock In \emph{Proceedings of the 21st International Conference on
  Computational Linguistics and the 44th annual meeting of the Association for
  Computational Linguistics (ACL-COLING)}, pages 729--736.

\bibitem[{Mori et~al.(1998)Mori, Tsuchiya, Yamaji, and Nagao}]{Mori1998Kana}
Shinsuke Mori, Masatoshi Tsuchiya, Osamu Yamaji, and Makoto Nagao. 1998.
\newblock \href {https://ci.nii.ac.jp/naid/110002917060/en/} {{Kana-Kanji}
  conversion by a stochastic model}.
\newblock \emph{Information Processing Society of Japan (IPSJ)}, pages
  2946--2953.

\bibitem[{Neubig et~al.(2013)Neubig, Watanabe, Mori, and
  Kawahara}]{Neubig2013Machine}
Graham Neubig, Taro Watanabe, Shinsuke Mori, and Tatsuya Kawahara. 2013.
\newblock \href {https://www.aclweb.org/anthology/P12-1018} {Machine
  translation without words through substring alignment}.
\newblock In \emph{Proceedings of the 50th Annual Meeting of the Association
  for Computational Linguistics (ACL)}, pages 165--174.

\bibitem[{Okuno and Mori(2012)}]{Okuno2012Using}
Yoh Okuno and Shinsuke Mori. 2012.
\newblock \href {https://www.aclweb.org/anthology/W12-4802} {An ensemble model
  of word-based and character-based models for {Japanese} and {Chinese} input
  method}.
\newblock In \emph{Proceedings of the Second Workshop on Advances in Text Input
  Methods}, pages 15--28.

\bibitem[{Verwimp et~al.(2017)Verwimp, Pelemans, Hamme, and
  Wambacq}]{Verwimp2017Character}
Lyan Verwimp, Joris Pelemans, Hugo~Van Hamme, and Patrick Wambacq. 2017.
\newblock \href {https://www.aclweb.org/anthology/E17-1040} {Character-word
  {LSTM} language models}.
\newblock In \emph{Proceedings of the 15th Conference of the European Chapter
  of the Association for Computational Linguistics (EACL)}, pages 417--427.

\bibitem[{Wang et~al.(2017{\natexlab{a}})Wang, Finch, Utiyama, and
  Sumita}]{wang-etal-2017-sentence}
Rui Wang, Andrew Finch, Masao Utiyama, and Eiichiro Sumita. 2017{\natexlab{a}}.
\newblock \href {https://doi.org/10.18653/v1/P17-2089} {Sentence embedding for
  neural machine translation domain adaptation}.
\newblock In \emph{Proceedings of the 55th Annual Meeting of the Association
  for Computational Linguistics (ACL)}, pages 560--566. Association for
  Computational Linguistics.

\bibitem[{{Wang} et~al.(2018){Wang}, {Utiyama}, {Finch}, {Liu}, {Chen}, and
  {Sumita}}]{8360031}
Rui {Wang}, Masao {Utiyama}, Andrew {Finch}, Lemao {Liu}, Kehai {Chen}, and
  Eiichiro {Sumita}. 2018.
\newblock \href {https://doi.org/10.1109/TASLP.2018.2837223} {Sentence
  selection and weighting for neural machine translation domain adaptation}.
\newblock \emph{IEEE/ACM Transactions on Audio, Speech, and Language
  Processing}, 26(10):1727--1741.

\bibitem[{Wang et~al.(2017{\natexlab{b}})Wang, Utiyama, Liu, Chen, and
  Sumita}]{wang-etal-2017-instance}
Rui Wang, Masao Utiyama, Lemao Liu, Kehai Chen, and Eiichiro Sumita.
  2017{\natexlab{b}}.
\newblock \href {https://doi.org/10.18653/v1/D17-1155} {Instance weighting for
  neural machine translation domain adaptation}.
\newblock In \emph{Proceedings of the 2017 Conference on Empirical Methods in
  Natural Language Processing (EMNLP)}, pages 1482--1488. Association for
  Computational Linguistics.

\bibitem[{Wang et~al.(2018)Wang, Utiyama, and Sumita}]{wang-etal-2018-dynamic}
Rui Wang, Masao Utiyama, and Eiichiro Sumita. 2018.
\newblock \href {https://www.aclweb.org/anthology/P18-2048} {Dynamic sentence
  sampling for efficient training of neural machine translation}.
\newblock In \emph{Proceedings of the 56th Annual Meeting of the Association
  for Computational Linguistics (ACL)}, pages 298--304.

\bibitem[{Wu et~al.(2018)Wu, Wu, Yang, Xu, Li, and Zhou}]{wu:2018}
Yu~Wu, Wei Wu, Dejian Yang, Can Xu, Zhoujun Li, and Ming Zhou. 2018.
\newblock \href
  {https://www.aaai.org/ocs/index.php/AAAI/AAAI18/paper/viewFile/16135/16117}
  {Neural response generation with dynamic vocabularies}.
\newblock In \emph{Thirty-Second AAAI Conference on Artificial Intelligence
  (AAAI)}, pages 5594--5601.

\bibitem[{Xiao et~al.(2019)Xiao, Li, Zhao, Wang, and Chen}]{xiao2019latt}
Fengshun Xiao, Jiangtong Li, Hai Zhao, Rui Wang, and Kehai Chen. 2019.
\newblock {Lattice-Based Transformer Encoder for Neural Machine Translation}.
\newblock In \emph{Proceedings of the 57th Annual Meeting of the Association
  for Computational Linguistics (ACL)}.

\bibitem[{Xiao et~al.(2008)Xiao, Liu, and WANG}]{Xiaoqiang1996A}
Jinghui Xiao, Bing~Quan Liu, and XiaoLong WANG. 2008.
\newblock \href {https://doi.org/10.3724/SP.J.1004.2008.00040} {A self-adaptive
  lexicon construction algorithm for {Chinese} language modeling}.
\newblock \emph{Acta Automatica Sinica}, 34(1):40--47.

\bibitem[{Yang et~al.(2012)Yang, Zhao, and Lu}]{Yang2012A}
Shaohua Yang, Hai Zhao, and Bao-liang Lu. 2012.
\newblock \href {https://www.aclweb.org/anthology/Y12-1036} {A machine
  translation approach for {Chinese} whole-sentence pinyin-to-character
  conversion}.
\newblock In \emph{Proceedings of the 26th Asian Pacific conference on language
  and information and computation (PACLIC)}, pages 333--342.

\bibitem[{Zhang et~al.(2017)Zhang, Wei, and Zhao}]{Zhang2017Tracing}
Xihu Zhang, Chu Wei, and Hai Zhao. 2017.
\newblock \href {https://arxiv.org/pdf/1712.04158.pdf} {Tracing a loose
  wordhood for {Chinese} input method engine}.
\newblock \emph{arXiv preprint arXiv:1712.04158}.

\bibitem[{Zhang et~al.(2018{\natexlab{a}})Zhang, Huang, and
  Zhao}]{zhang2018SubMRC}
Zhuosheng Zhang, Yafang Huang, and Hai Zhao. 2018{\natexlab{a}}.
\newblock \href {https://www.aclweb.org/anthology/C18-1153} {Subword-augmented
  embedding for cloze reading comprehension}.
\newblock In \emph{Proceedings of the 27th International Conference on
  Computational Linguistics (COLING)}, pages 1802--1814.

\bibitem[{Zhang et~al.(2018{\natexlab{b}})Zhang, Huang, Zhu, and
  Zhao}]{zhang2018char}
Zhuosheng Zhang, Yafang Huang, Pengfei Zhu, and Hai Zhao. 2018{\natexlab{b}}.
\newblock \href {https://arxiv.org/pdf/1808.02772.pdf} {Effective
  character-augmented word embedding for machine reading comprehension}.
\newblock In \emph{Proceedings of the Seventh CCF International Conference on
  Natural Language Processing and Chinese Computing (NLPCC)}, pages 27--39.

\bibitem[{Zhang et~al.(2018{\natexlab{c}})Zhang, Li, Zhu, and
  Zhao}]{zhang2018DUA}
Zhuosheng Zhang, Jiangtong Li, Pengfei Zhu, and Hai Zhao. 2018{\natexlab{c}}.
\newblock \href {https://www.aclweb.org/anthology/C18-1317} {Modeling
  multi-turn conversation with deep utterance aggregation}.
\newblock In \emph{Proceedings of the 27th International Conference on
  Computational Linguistics (COLING)}, pages 3740--3752.

\bibitem[{Zhang and Zhao(2018)}]{zhang2018OneShot}
Zhuosheng Zhang and Hai Zhao. 2018.
\newblock \href {https://www.aclweb.org/anthology/C18-1038} {One-shot learning
  for question-answering in gaokao history challenge}.
\newblock In \emph{Proceedings of the 27th International Conference on
  Computational Linguistics (COLING)}, pages 449--461.

\bibitem[{Zhao et~al.(2006)Zhao, Huang, Li, and Kudo}]{Hai2006An}
Hai Zhao, Chang-Ning Huang, Mu~Li, and Taku Kudo. 2006.
\newblock \href {https://www.aclweb.org/anthology/W06-0127} {An improved
  {Chinese} word segmentation system with conditional random field}.
\newblock In \emph{Proceedings of the Fifth SIGHAN Workshop on Chinese Language
  Processing}, pages 162--165.

\bibitem[{Zhou et~al.(2016)Zhou, Cao, Wang, Li, and Xu}]{zhou:2016}
Jie Zhou, Ying Cao, Xuguang Wang, Peng Li, and Wei Xu. 2016.
\newblock \href {https://doi.org/10.1162/tacl_a_00105} {Deep recurrent models
  with fast-forward connections for neural machine translation}.
\newblock \emph{Transactions of the Association for Computational Linguistics
  (TACL)}, 4:371--383.

\bibitem[{Zhou and Zhao(2019)}]{zhou-2019-HPSG}
Junru Zhou and Hai Zhao. 2019.
\newblock {Head-Driven Phrase Structure Grammar Parsing on Penn Treebank}.
\newblock In \emph{Proceedings of the 57th Annual Meeting of the Association
  for Computational Linguistics (ACL)}.

\bibitem[{Zhu et~al.(2018)Zhu, Zhang, Li, Huang, and Zhao}]{Zhu2018lingke}
Pengfei Zhu, Zhuosheng Zhang, Jiangtong Li, Yafang Huang, and Hai Zhao. 2018.
\newblock \href {https://www.aclweb.org/anthology/C18-2024} {Lingke: A
  fine-grained multi-turn chatbot for customer service}.
\newblock In \emph{Proceedings of the 27th International Conference on
  Computational Linguistics (COLING), System Demonstrations}, pages 108--112.

\end{thebibliography}
\bibliographystyle{acl_natbib}

\end{document}